\documentclass[runningheads,a4paper]{llncs}

\usepackage{amssymb}
\usepackage{graphicx}

\usepackage{enumitem}
\def\Tref#1{Table~\ref{#1}}
\def\equo#1{``#1''}
\def\notion#1{{\emph{#1}}}
\def\text#1{{\emph{#1}}}
\def\abbrdef#1{(#1)}
\newcommand{\argmax}{\mathop{\mathrm{argmax}}}

\usepackage{url}  
\newcommand{\keywords}[1]{\par\addvspace\baselineskip
\noindent\keywordname\enspace\ignorespaces#1}

\begin{document}

\mainmatter  

\title{SubGram: Extending Skip-gram Word Representation with Substrings}

\titlerunning{SubGram: Extending Skip-gram Word Representation with Substrings}

\author{Tom Kocmi \and Ond{\v r}ej Bojar}
\authorrunning{SubGram: Extending Skip-gram Word Representation with Substrings}

\institute{Charles University in Prague, \\
	    Faculty of Mathematics and Physics,\\
Institute of Formal and Applied Linguistics\\
{\tt \{kocmi,bojar\}@ufal.mff.cuni.cz}}

\maketitle

\begin{abstract}
Skip-gram (word2vec) is a recent method for creating vector representations of
words (\equo{distributed word representations}) using a neural
network. The representation gained popularity in various areas of
natural language processing, because it
seems to capture syntactic and semantic information
about words without any explicit supervision in this respect.
%

We propose SubGram, a refinement of the Skip-gram model to consider
also the word structure during the training process, achieving large gains 
on the Skip-gram original test set.

\end{abstract}

\section{Introduction}

Vector representations of words learned using neural networks
\abbrdef{NN} have proven helpful in many
algorithms for image annotation
\cite{Lazaridou2015} or \cite{Weston2011}, language modeling
\cite{Schwenk2004}, \cite{Schwenk2006} and \cite{Mnih2007} or other natural language processing
\abbrdef{NLP} tasks
\cite{Soricut2015} or \cite{Wang2014}.

Traditionally, every input word
of an NN is stored in the
\equo{one-hot} representation, where the vector has only one element set to one and the 
rest of the vector are zeros. The size of the vector equals to the size of the vocabulary.
The NN is trained to perform some prediction, e.g. to predict surrounding words
given a word of interest. Instead of using this prediction capacity in some
task, the practice is to extract the output of NN's hidden layer of each word (called \notion{distributed representation}) and directly use this deterministic
mapping $vec(\cdot)$ of word forms to the vectors of real numbers as the word representation.

The input one-hot representation of words has two weaknesses: the bloat of the size of 
the vector with more words in vocabulary and the inability to
provide any explicit semantic or syntactic information to the NN.

The learned distributed representation of words relies on much shorter vectors
(e.g. vocabularies containing millions words are represented in vectors of a few hundred
elements) and semantic or syntactic information is often found to be implicitly
present (\equo{embedded}) in the vector space.
%
For example, the Euclidean distance between two words in the
vector space may be related to semantic or syntactic similarity between them.

\subsection{Skip-gram Model}

The authors of \cite{Mikolov2013} created a model
called Skip-gram, in which linear vector operations allow to find related words
with surprisingly good results. For instance
$vec(\text{king}) - vec(\text{man}) 
+ vec(\text{woman})$ gives a value close to $vec(\text{queen})$.

In this paper, we extend Skip-gram model with the internal word
structure and show how it improves the performance on embedding morpho-syntactic information.

The Skip-gram model defined in \cite{Mikolov2013} is trained to predict context words of the
input word.  Given a corpus $T$ of words $w$ and their context words $c \in
C(w)$ (i.e. individual words $c$ appearing close the original word $w$), it
considers the conditional probabilities $p(c|w)$. The training finds the 
parameters $\theta$ of $p(c|w; \theta)$ to maximize the corpus probability:

\begin{equation}
\argmax_{\theta} \prod_{w \in T} \prod_{c \in C(w)}p(c|w;\theta)
\end{equation}

The Skip-gram model is a classic NN, where activation functions are removed and hierarchical 
soft-max \cite{Morin2005} is used instead of soft-max normalization. 
The input representation 
is one-hot so the activation function is not needed on hidden
layer, there is nothing to be summed up. This way, the model is learned much faster than comparable
non-linear NNs and lends itself to linear vector operations possibly useful for
finding semantically or syntactically related words.

\section{Related Work}

In \cite{Qiu2014} was proposed to append part-of-speech \abbrdef{POS} tags to each
word and train Skip-gram model on the new vocabulary. This avoided
conflating, e.g. nouns and verbs, leading to a better performance,
at the cost of
(1) the reliance on POS tags and their accurate estimation
and (2) the increased sparsity of the data due to the
larger vocabulary.

The authors in \cite{Yoon2015} used character-level input to train language models using a
complex setup of NNs of several types.
Their model was able to assign meaning to out-of-vocabulary 
words based on the closest neighbor. One disadvantage of the model 
is its need to run the computation on a GPU for a long time.

The authors of \cite{Cui2014} proposed an extension of Skip-gram model which uses character 
similarity of words to improve performance on syntactic and semantic tasks. 
They are using a set of similar words as additional features for the NN.
Various similarity measures are tested: Levenshtein, longest common substring, morpheme 
and syllable similarity.

The authors of \cite{Bian2014} added the information about word's root, 
affixes, syllables, synonyms, antonyms and POS tags to continuous bag-of-words model
\abbrdef{CBOW} proposed by \cite{Mikolov2013} and showed how these types of knowledge lead to better word embeddings. 
The CBOW model is a simpler model with usually worse performance than Skip-gram.


\section{SubGram}

We propose a substring-oriented extension of Skip-gram model which induces vector embeddings
from character-level structure of individual words. This approach gives the
NN more information about the examined word with no drawbacks
in data sparsity or reliance on explicit linguistic annotation.
 
We append the characters \^{} and \$ to the word to indicate its beginning and end. In order to generate the vector of substrings, we take all character bigrams, trigrams
etc. up to the length of the word. This way, even the word itself is represented
as one of the substrings. For the NN, each input word is then represented as a binary vector 
indicating which substrings appear in the word.

The original Skip-gram model \cite{Mikolov2013} uses one-hot
representation of a word in vocabulary as the input vector. This representation
makes training fast because no summation or normalization is needed. The weights
$w_i$ of the input word $i$ can be directly used
as the output of hidden layer $h$ (and as the distributed word
representation): $h_j = w_ij$

In our approach, we provide the network with a binary vector representing all
substrings of the word. To compute the input of hidden layer we decided to use mean value as it is computationally simpler than sigmoid:

\begin{equation}
h_j = \frac{\sum_{i=1}^{|X|}x_i*w_{ij}}{|S|}
\end{equation}

\noindent
where $|S|$ is the number of substrings of the word $x$.

\section{Evaluation and Data Sets}

We train our NN on words and their contexts extracted from the English wikipedia dump from May 2015. We have cleaned the data by replacing all numbers with 0 and removing special characters except those usually present in the English text like dots, brackets, apostrophes etc. For the final training data we have randomly selected only 2.5M segments (mostly sentences). It consist of 96M running words with the vocabulary size of 1.09M distinct
word forms. 

We consider only the 141K most frequent word forms to simplify the
training.
The remaining word forms fall out of vocabulary (OOV), so the original
Skip-gram cannot provide them with any vector representation. Our SubGram relies
on known substrings and always provides at least some approximation.

We test our model on the original test set \cite{Mikolov2013}. The test set
consists of 19544 \equo{questions}, of which 8869 are called \equo{semantic} and
10675 are called \equo{syntactic} and further divided into 14 types, see Table \ref{samples}.
Each question contains two pairs of words ($x_1, x_2,
y_1, y_2$) and captures relations like ``What is to `woman' ($y_1$) as `king'
($x_2$) is to
`man' ($x_1$)?'', together with the expected answer `queen' ($y_2$). The model
is evaluated by finding the
word whose representation is the nearest (cosine similarity) to the vector $vec(\text{king}) - vec(\text{man}) +
vec(\text{woman})$.
If the nearest neighbor is $vec(\text{queen})$, we consider the question answered correctly.

\begin{table}[t]
\begin{center}
\begin{tabular}{lr@{~--~}l}
Question Type & \multicolumn{2}{c}{Sample Pair} \\
\hline
capital-countries &   Athens & Greece \\
capital-world &  Abuja & Nigeria \\
currency &  Algeria & dinar\\
city-in-state &  Houston & Texas \\
family & boy & girl  \\
\hline
adjective-to-adverb & calm & calmly \\
opposite &  aware & unaware \\
comparative &  bad & worse\\
superlative &  bad & worst\\
present-participle &  code & coding\\
nationality-adjective &  Albania & Albanian\\
past-tense & dancing & danced \\
plural &  banana & bananas\\
plural-verbs &  decrease & decreases\\
\end{tabular}
\end{center}
\caption{\label{samples}Mikolov's test set question types, the upper part are \equo{semantic}
questions, the lower part are \equo{syntactic}.}
\end{table}

In this work, we use Mikolov's test set which is used in many
papers. After a closer examination we came to the conclusion, that it does not test what
the broad terms \equo{syntactic} and \equo{semantic relations} suggest.
\equo{Semantics} is covered by questions of only 3 types: 
predict a city based on a country or state, currency name
from the country and the feminine variant of nouns denoting family
relations. The authors of \cite{Vylomova2015} showed, that many other
semantic relationships could be tested, e.g. walk-run, dog-puppy,
bark-dog, cook-eat and others.

\equo{Syntactic} questions cover a wider range of relations at the boundary of
morphology and syntax. The problem is that all questions of a given type are
constructed from just a few dozens of word pairs, comparing pairs with each
other. Overall, there are 313 distinct pairs throughout the whole syntactic
test set of 10675 questions, 
which means only around 35 different pairs per question set.
Moreover, of the 313 pairs, 286 pairs are regularly formed (e.g. by adding the
suffix `ly' to change an adjective into the corresponding adverb). Though it has to be mentioned that original model could not use this kind of information.

We find such a small test set unreliable to answer the question
whether the embedding captures semantic and syntactic properties of words.

\subsection{Rule-Based Baseline Approach}

Although the original test set has been used to compare
results in several papers, no-one tried to process it with some baseline approach. Therefore, we
created a very simple set of rules for comparison on the syntactic part of the test set. The rules cover only the most frequent grammatical phenomenona.


\begin{itemize}[noitemsep]
\item adjective-to-adverb: Add \text{ly} at the end of the adjective.
\item opposite: Add \text{un} at the beginning of positive form.
\item comparative: If the adjective ends with \text{y}, replace it with
\text{ier}. If it ends with \text{e}, add \text{r}. Otherwise add \text{er} at the end.
\item superlative: If the adjective ends with \text{y}, replace it with
\text{iest}. If it ends with \text{e}, add \text{st}. Otherwise add \text{est}
at the end.
\item present-participle: If the verb ends with \text{e}, replace it with
\text{ing}, otherwise add \text{ing} at the end.
\item nationality-adjective: Add \text{n} at the end, e.g. \text{Russia}
$\rightarrow$ \text{Russian}.
\item past-tense: Remove \text{ing} and add \text{ed} at the end of the verb.
\item plural: Add \text{s} at the end of the word.
\item plural-verbs: If the word ends with a vowel, add \text{es} at the end, else add \text{s}.
\end{itemize}

\subsection{Our Test Set}

We have decided to create more general test set which would consider more than 35 pairs per question set. Since we are interested
in morphosyntactic relations, we extended only the questions of the
\equo{syntactic} type with exception of nationality adjectives which is already covered completely in original test set.

We constructed the pairs more or less manually, taking
inspiration in the Czech side of the CzEng corpus \cite{czeng10:lrec2012}, where explicit morphological
annotation allows to identify various pairs of Czech words (different grades of
adjectives, words and their negations, etc.). The word-aligned English words
often shared the same properties. Another sources of pairs were acquired from
various webpages usually written for learners of English. For example for verb tense, we relied on a freely
available list of English
verbs and their morphological variations. We have included 100--1000 different
pairs for each question set. The questions were constructed from the
pairs similarly as by Mikolov: generating all possible pairs of pairs. This
leads to millions of questions, so we randomly selected 1000 instances per
question set, to keep the test set in the same order of magnitude.
Additionally, we decided to extend set of questions on opposites to cover not
only opposites of adjectives but also of nouns and verbs.

In order to test our extension of Skip-gram on out-of-vocabulary words, we
created an additional subset of our test set with questions where at least one of
$x_1, x_2$ and $y_1$ is not among the known word forms. Note that the last word $y_2$ must be in vocabulary in order to check if the output vector is correct.


\section{Experiments and Results}

We used a Python implementation of  
word2vec\footnote{http://radimrehurek.com/gensim \newline
Gensim implements the model twice, in Python and an optimized version in C. For
our prototype, we opted to modify the Python version, which unfortunately
resulted in a code about 100 times slower and forced us to train the model
only on the 96M word corpus as opposed to Mikolov's 100000M
word2vec training data used in training of the released model.} as the basis for our
SubGram, which we have made freely available \footnote{https://github.com/tomkocmi/SubGram}.

We limit the vocabulary, requiring each word form to appear
at
least 10 times in the corpus and each substring to appear at least 500 times in
the corpus.
This way, we get the 141K unique words mentioned above and 170K unique substrings (+141K words, as we downsample words separately). 

Our word vectors have the size of 100. The size of the context window is 5.

The accuracy is computed as the number of correctly answered questions divided
by the total number of questions in the set. Because the Skip-gram cannot answer
questions containing OOV words, we also provide results with such questions
excluded from the test set (scores in brackets).

\begin{table}
\begin{center}
\small
\begin{tabular}{lr|rl|rl|rl}
 & \llap{\bf Rule based} & \multicolumn{2}{c|}{\bf Released Skip-gram} & \multicolumn{2}{c|}{\bf Our Skip-gram} & \multicolumn{2}{c}{\bf SubGram} \\
capital-countries    & 0\% & 18.6\% & (24.7\%) & 71.9\% &  (71.9\%)  & 0\% & (0\%)  \\
capital-world        & 0\% &  2.2\% & (15.0\%) & 53.6\% &  (54.6\%)   & 0\% & (0\%) \\
currency             & 0\% &    7\% & (12.2\%) & 3\% &  (4.7\%)  & 0.1\% & (0.2\%) \\
city-in-state        & 0\% &  9.2\% &   (14\%) & 40.5\% &  (40.5\%)  & 0.1\% & (0.1\%) \\
family               & 0\% & 84.6\% & (84.6\%) & 82.6\% & (82.6\%)   & 5.9\% & (5.9\%)  \\
\bf Overall semantic & 0\% & 10.2\% & (24.8\%) & 47.7\% & (50\%)   & 0\% & (0\%)  \\ \hline
adjective-to-adverb   & 90.6\% & 28.5\% & (28.5\%) & 16.3\% & (16.3\%)   & 73.7\% & (73.7\%) \\
opposite              & 65.5\% & 42.7\% & (42.7\%) &  9.4\% & (10.1\%)  & 43.1\% & (46.3\%) \\
comparative           & 89.2\% & 90.8\% & (90.8\%) &  72.1\% & (72.1\%)  & 46.5\% & (46.5\%) \\
superlative           & 88.2\% & 87.3\% & (87.3\%) & 24.4\% & (25.9\%)  & 45.9\% & (48.8\%)  \\
present-participle    & 87.9\% & 78.1\% & (78.1\%) & 44.2\% & (44.2\%)  & 43.5\% & (43.5\%) \\
nationality-adjective & 31.7\% & 13.3\% & (21.9\%) & 60.4\% & (60.4\%)  & 21.8\% & (21.8\%) \\
past-tense            & 42.5\% &   66\% &   (66\%) & 35.6\% & (35.6\%)  & 15.8\% & (15.8\%) \\
plural                & 86.5\% & 89.9\% & (89.9\%) & 46.8\% & (46.8\%)  & 44.7\% & (44.7\%) \\
plural-verbs          & 93.3\% & 67.9\% & (67.9\%) & 51.5\% & (51.5\%)  & 74.3\% & (74.3\%) \\
\bf Overall syntactic & 71.9\% & 62.5\% & (66.5\%) & 42.5\% & (43\%)   & 42.3\% & (42.7\%)  \\
\end{tabular}
\end{center}
\caption{\label{results-table} Results on original test set questions.
The values in brackets are based on questions without any OOVs.}
\end{table}

\begin{table}
\begin{center}
\small
\begin{tabular}{lr|rl|rl|rl|r}
 & \llap{\bf Rule based} & \multicolumn{2}{c|}{\bf Released Skip-gram} & \multicolumn{2}{c|}{\bf Our Skip-gram} & \multicolumn{3}{c}{\bf SubGram } \\
\bf Type & \multicolumn{7}{c|}{\bf Our test set} & \bf OOV\\ \hline
adjective-to-adverb  & 68.4\% & 18.8\% & (20.9\%) & 1.9\%&  (3.7\%)  & 32.3\% & (62.7\%)  & 2.3\%\\
opposite             &  3.7\% & 6.3\% & (6.4\%)  & 5.3\%&  (5.6\%) & 0.6\% &  (0.6\%) & 0.7\%\\
comparative          & 90.2\% & 67.1\% & (68.9\%)& 12\% &  (31.5\%)  & 14.4\% &  (37.8\%) & 0\%\\
superlative          & 92.5\% & 57\% & (59.9\%) & 4.4\%&  (16.1\%)  & 12.2\% & (44.7\%)  & 0.5\%\\
present-participle   & 88.7\% & 50.2\% & (53\%) & 12.8\%&  (16.2\%)  & 37.3\% & (47.3\%)  & 4.8\%\\
past-tense            &  75\% & 53.5\% & (56.5\%) & 17.1\%&  (22.8\%)  & 24.2\% & (32.3\%)  & 0.5\%\\
plural               & 26.8\% & 39.1\% & (42.1\%) & 8.9\%&  (13.8\%)  & 13.6\% & (21.1\%)  & 1.7\%\\
plural-verbs          & 85.8\% & 56.1\% & (59\%) &15.4\%&  (20.3\%)  & 44.3\% & (58.5\%)  & 2\%\\
\bf Overall syntactic & 66.4\% & 43.5\% & (45.9\%) & 9.7\%&  (15.4\%)  & 22.4\% & (35.4\%)  & 1.6\%\\
\end{tabular}
\end{center}
\caption{\label{results-table2} Results on our test set questions.}
\end{table}

\Tref{results-table} and \Tref{results-table2} report the results. The first
column shows the rule-based approach. The column \equo{Released Skip-gram} shows
results of the model released by
Mikolov\footnote{https://code.google.com/archive/p/word2vec/} and was trained on
a 100 billion word corpus from Google News and generates 300 dimensional vector
representation. The third column shows Skip-gram model trained on our training data,
the same data as used for the training of the SubGram. Last column shows
the results obtained from our SubGram model.

Comparing Skip-gram and SubGram on the original test set (\Tref{results-table}),
we see that our SubGram outperforms Skip-gram in several morpho-syntactic
question sets but over all performs similarly (42.5\% vs. 42.3\%). On
the other hand, it does not capture the tested semantic relations at all,
getting a zero score on average. 

When comparing models on our test set (\Tref{results-table2}), we see that
given the same training set, SubGram significantly outperforms Skip-gram model
(22.4\% vs. 9.7\%).  The performance of Skip-gram trained on the much larger
dataset is higher (43.5\%) and it would be interesting to see the SubGram model,
if we could get access to such training data. Note however, that the Rule-based
baseline is significantly better on both test sets.

The last column suggests that the performance of our model on OOV words is not
very high, but it is still an improvement over flat zero of the 
Skip-gram model. The performance on OOVs is expected to be lower, since the
model has no knowledge of exceptions and can only benefit from regularities in
substrings.

\section{Future Work}

We are working on a better test set for word embeddings which would include many
more relations over a larger vocabulary especially semantics relations. We want to extend the test set with Czech and perhaps other
languages, to see what word embeddings can bring to languages morphologically
richer than English.

As shown in the results, the rule based approach outperform NN approach on this type of task, therefore we would like to create a hybrid system which could use rules and part-of-speech tags. We will also include morphological tags in the model as proposed in  \cite{Qiu2014} 
but without making the data sparse.

Finally, we plan to reimplement SubGram to scale up to larger training data. 

\section{Conclusion}

We described SubGram, an extension of the Skip-gram model 
that considers also substrings of input words. The learned embeddings then better capture almost all
morpho-syntactic relations tested on test set which we extended from original described in \cite{Mikolov2013}. This test set is released for the public use\footnote{https://ufal.mff.cuni.cz/tom-kocmi/syntactic-questions}.

An useful feature of our model is the ability to generate vector
embeddings even
for unseen words. This could be exploited by NNs also in different tasks.

%

\section{Acknowledgment}
This work has received funding from the European Union's Horizon 2020 research and innovation programme under grant agreement no. 645452 (QT21), the grant GAUK 8502/2016, and SVV project number 260 333.

This work has been using language resources developed,
stored and distributed by the LINDAT/\discretionary{}{}{}CLARIN project of the
Ministry of Education, Youth and Sports of the Czech Republic (project LM2015071).

\bibliography{biblio}
\bibliographystyle{splncs}

\end{document}